%% file: arxiv_version.tex
\documentclass[runningheads]{llncs}
\usepackage[T1]{fontenc}
\usepackage{graphicx,verbatim}
\usepackage{graphicx}
\usepackage{comment}
\usepackage{amsfonts}
\usepackage{amsmath}
\usepackage{array}
\usepackage{color}
\usepackage{marvosym} 
\definecolor{myred}{rgb}{.8,.0,.0}

\begin{document}

\title{Med-DualLoRA: Local Adaptation of Foundation Models for 3D Cardiac MRI}

\author{Joan Perramon-Llussà\inst{1}\textsuperscript{\Letter} \and
Amelia Jiménez-Sánchez\inst{1} \and
Grzegorz Skorupko\inst{1} \and
Fotis Avgoustidis\inst{1} \and
Carlos Martín-Isla\inst{1} \and
Karim Lekadir\inst{1,2} \and
Polyxeni Gkontra\inst{1}
}

\authorrunning{Perramon-Llussà et al.}

\institute{
Artificial Intelligence in Medicine Lab (BCN-AIM), Departament de Matemàtiques i Informàtica, Universitat de Barcelona, Spain\\
\email{\Letter\ joan.perramon@ub.edu}
\and
Institució Catalana de Recerca i Estudis Avançats (ICREA), Barcelona, Spain
}

\maketitle
\begin{abstract}
Foundation models (FMs) show great promise for robust downstream performance across medical imaging tasks and modalities, including cardiac magnetic resonance (CMR), following task-specific adaptation. However, adaptation using single-site data may lead to suboptimal performance and increased model bias, while centralized fine-tuning on clinical data is often infeasible due to privacy constraints. Federated fine-tuning offers a privacy-preserving alternative; yet conventional approaches struggle under heterogeneous, non-IID multi-center data and incur substantial communication overhead when adapting large models. In this work, we study federated FM fine-tuning for 3D CMR disease detection and propose Med-DualLoRA, a client-aware parameter-efficient fine-tuning (PEFT) federated framework that disentangles globally shared and local low-rank adaptations (LoRA) through additive decomposition. Global and local LoRA modules are trained locally, but only the global component is shared and aggregated across sites, keeping local adapters private. This design improves personalization while significantly reducing communication cost, and experiments show that adapting only two transformer blocks preserves performance while further improving efficiency. We evaluate our method on a multi-center state-of-the-art cine 3D CMR FM fine-tuned for disease detection using ACDC and combined M\&Ms datasets, treating each vendor as a federated client. Med-DualLoRA achieves statistically significant improved performance (balanced accuracy 0.768, specificity 0.612) compared to other federated PEFT baselines, while maintaining communication efficiency. Our approach provides a scalable solution for local federated adaptation of medical FMs under realistic clinical constraints.


\end{abstract}
\section{Introduction}
Cardiac magnetic resonance (CMR) is the reference modality for the assessment of cardiac heart structure and function, providing rich structural and functional information for diagnosis, risk stratification, and longitudinal monitoring~\cite{ref_pennell_cmr}. Although deep learning has demonstrated strong potential in automating CMR-based tasks~\cite{ref_litjens}, most models are trained on data from a single institution and, therefore, exhibit limited generalization to external hospitals due to domain shift, scanner heterogeneity, and annotation variability~\cite{ref_guan,ref_zech}. Foundation models (FMs) trained on large-scale heterogeneous datasets have recently demonstrated improved transferability across medical imaging tasks and modalities, including CMR, following task-specific adaptation~\cite{ref_azizi,ref_jacob_fm_cmr,ref_wang_triad}. However, site-specific adaptation alone can fail to yield optimal performance and can increase model bias~\cite{ref_bommasani}, while centralized fine-tuning using clinical data is often impractical due to privacy constrains~\cite{ref_rieke,ref_jimenez2023memory}. 

To address these challenges, federated fine-tuning of FMs provides a promising privacy-preserving strategy for multi-site adaptation~\cite{ref_mouheb}. Nonetheless, federation of large FMs involves repeated transmission of high-dimensional parameter updates, resulting in substantial communication and computational overhead that can be prohibitive in privacy-sensitive and resource-constrained clinical environments~\cite{ref_rieke}. Parameter-efficient fine-tuning (PEFT) techniques aim to adapt pretrained models while updating only a small subset of parameters. Linear probing represents the simplest PEFT strategy by freezing the backbone and training a lightweight task-specific head~\cite{truong2021transferable}, whereas Low-Rank Adaptation (LoRA) injects low-rank adapters into pretrained models while freezing backbone weights~\cite{ref_hu_lora}. By reducing trainable parameters, these approaches improve communication efficiency in federated learning (FL)~\cite{ref_mouheb}. However, aggregating client-specific adapters into a single global model under heterogeneous (non-IID) data distributions can bias the model toward dominant clients, potentially leading to suboptimal performance on minority or distribution-shifted centers~\cite{ref_zhao}. 

To tackle this limitation, client-aware FL methods introduce client-specific components to better capture local characteristics. In the natural language processing domain, FedALoRA adaptively initializes local models before each training iteration by fusing global and client parameters to balance general and domain-specific knowledge~\cite{ref_Yi2025}. FDLoRA proposes dual LoRA modules that decouple global and local adaptations with adaptive fusion in large language models~\cite{ref_qi_fedlora}. The concept of decoupling global and local representations was previously explored in deep feed-forward architectures for 2D imaging, employing base and personalization layers to address heterogeneity in non-medical datasets~\cite{ref_arivazhagan_personal_fl}. Despite these advances, existing approaches in 3D medical imaging either rely on implicit regularization or modify aggregation heuristics~\cite{ref_he_ratemylora}, without explicitly disentangling globally transferable and client-specific low-rank adaptations.

To bridge this gap, we propose Med-DualLoRA, a client-aware and communication -efficient federated fine-tuning framework for 3D CMR FMs that explicitly disentangles shared and local low-rank adaptations through an additive formulation. By separating globally transferable and site-specific representations, Med-DualLoRA improves robustness under distribution shift while maintaining efficient communication. Our contributions are summarized as follows:
\begin{enumerate}
\item We present the first in-depth analysis of key fine-tuning strategies for 3D CMR FMs in multi-center settings, using CineMA~\cite{ref_fu_cinema} as a case study.

\item We introduce Med-DualLoRA, a federated fine-tuning method for medical imaging FMs that learns global and client-specific adaptations separately. To the best of our knowledge, this is the first work to introduce such structural separation for federated adaptation of 3D medical imaging FMs.

\item We validate the proposed framework on multi-center CMR disease detection across two publicly available MICCAI challenge datasets (ACDC and M\&Ms), demonstrating statistically significant improved cross-center generalization and per-site robustness over federated PEFT baselines.

\item We show that applying Med-DualLoRA to only two FM transformer blocks preserves performance while substantially reducing communication cost.
\end{enumerate}
\input{figures/figure_federated_dual_lora}

\section{Med-DualLora Framework}
Inspired by LoRA, we inject trainable low-rank adapters into the linear projections of the attention layers of a frozen state-of-the-art CMR pretrained FM. These adapters augment the original transformations through additive low-rank updates rather than replacing the backbone weights, preserving the pretrained representations while introducing only a small number of additional parameters. An overview of the proposed Med-DualLoRA architecture is provided in Fig.~\ref{fig:dual_lora_architecture}b.

\paragraph{\textbf{Federated Dual Low-Rank Adaptation.}}  
For a linear projection \(W \in \mathbb{R}^{d \times m}\) and input \(x \in \mathbb{R}^{m}\), being \(m\) the input dimensionality and \(d\) the output dimensionality of the linear projection, we replace it with a dual LoRA module. For client \(i\), the forward pass is:

\begin{equation}
    f_i(x) = W x 
    + \frac{\alpha}{r} B_{g} A_{g} x
    + \frac{\alpha}{r} B_{l_i} A_{l_i} x,\label{eq:dual_lora}
\end{equation}

where \(A_g, B_g \in \mathbb{R}^{r \times m}, \mathbb{R}^{d \times r}\) denote the shared (global) LoRA parameters, and \(A_{l_i}, B_{l_i} \in \mathbb{R}^{r \times m}, \mathbb{R}^{d \times r}\) denote the client-specific (local) LoRA parameters. Here, \(r\) is the rank of the adaptation and \(\alpha\) is a scaling factor. During local training, both global and local parameters are trainable. We keep the backbone weights frozen. The total number of trainable parameters per layer is \(2r(d + m)\), which is orders of magnitude smaller than full fine-tuning.

\paragraph{\textbf{Federated Optimization.}}  
Each client optimizes both global and local LoRA parameters on its own dataset. After local training, only the global parameters are transmitted and aggregated using weighted averaging:

\begin{equation}
    A_g \leftarrow \sum_{i=1}^{N} \frac{n_i}{n} A_{g_i}, 
    \qquad
    B_{g} \leftarrow \sum_{i=1}^{N} \frac{n_i}{n} B_{g_i},
\end{equation}

where \(n_i\) is the number of samples of client \(i\) and \(n = \sum_i n_i\). The local parameters \((A_{l_i}, B_{l_i})\) remain strictly private and are never shared. This separation enables collaborative learning of globally transferable representations with reduced communication while capturing client-specific variations under heterogeneous distributions.

\section{Experimental Setup}

\paragraph{\textbf{Datasets.}}

We fine-tune and evaluate our method on two publicly available cine cardiac MRI benchmarks: the Automated Cardiac Diagnosis Challenge (ACDC)~\cite{ref_bernard2018acdc} and a curated combined version of the Multi-centre, Multi-vendor \& Multi-disease (M\&Ms-1 and M\&Ms-2) datasets~\cite{ref_campello2021mnms,ref_martin_isla2023mnms}. The combined M\&Ms collection follows the curation described in~\cite{ref_skorupko_fednnunet}. ACDC comprises 150 patients acquired under controlled conditions, whereas the merged M\&Ms dataset contains 543 multi-center, multi-vendor acquisitions from five institutions and four vendors (Siemens, GE, Philips, Canon), reflecting realistic clinical variability in scanner models, field strengths, slice thickness, and acquisition protocols.

Images are preprocessed following the original CineMA pipeline~\cite{ref_fu_cinema}. Only SAX images of the end-diastole and end-systole frames are used for binary disease detection. We use the original ACDC dataset train/val/test splits. For the combined M\&Ms dataset, we retain the split assignments from M\&Ms2 where available, and use the corresponding M\&Ms1 splits for images not included in M\&Ms2. Vendors and ACDC are treated as separate federated clients. Class distributions are reported in Table~\ref{tab:data_distribution}.

\input{tables/table_data_distribution}

\paragraph{\textbf{Baselines.}}
 While our method is model agnostic, we choose CineMA FM~\cite{ref_fu_cinema} as the pretrained backbone for our experiments, given its strong performance on diverse clinical CMR downstream tasks. We compare our federated LoRA method against three data centralized and two data FL baselines (Fig.~\ref{fig:dual_lora_architecture}a). Centralized full fine-tuning trains the entire backbone and head on pooled data, providing an upper-bound performance reference. Linear probing, i.e. head-only tuning, in both centralized and FL settings, updates only the classification head while keeping the backbone frozen, representing minimal communication. Standard LoRA centralized and FL fine-tunes low-rank adapters in select transformer blocks, illustrating the trade-off between trainable parameters and performance. We also compare our method to client-specific models trained only with their private data to have a local baseline (Fig.~\ref{fig:performance-pareto}a).

\paragraph{\textbf{Federated Training Protocol.}}
At initialization, the central server distributes the pretrained backbone to all participating clients. During each communication round, clients perform local optimization using their private data, updating only the designated trainable components (classification head, LoRA adapters, or dual LoRA adapters), while the backbone remains frozen. The resulting updates are transmitted to the server and aggregated using FedAvg~\cite{ref_mcmahan2017fedavg}. For the Med-DualLoRA scenario, only the global LoRA adapters are shared and aggregated across clients, while keeping the local LoRA on-site. The aggregated parameters are then redistributed to the clients for the next training round and added to local LoRA modules as in Equation~\ref{eq:dual_lora} (Fig.\ref{fig:dual_lora_architecture}b). At inference time, for Med-DualLoRA, each client loads the common global LoRA and their private local LoRA into transformer layers and combines them in the same manner.

\paragraph{\textbf{Implementation Details.}}
All models are implemented in PyTorch. LoRA adapters are injected into the attention and MLP projections of selected transformer blocks. Based on validation experiments with different LoRA ranks and scaling factors, we select $r=4$ and $\alpha=8$ as it provides the best trade-off between communication efficiency and performance. For Med-DualLoRA, global and local adapters share the same rank, scaling factor, and target modules. Models are fine-tuned using the cross-entropy loss for binary classification. Optimization is performed using AdamW with weight decay, with learning rates selected per configuration based on dataset size and training setting. Centralized models are trained for up to 500 epochs with early stopping based on validation loss, while in the FL setting each client performs one local epoch per communication round using the same stopping criterion. Centralized experiments are conducted on a single NVIDIA GeForce RTX 4090 GPU, while federated experiments run on a dual NVIDIA L40S system to emulate concurrent client updates. All experiments are run for three random seeds and averaged across seeds. We make the code publicly available at https://github.com/joanperramon/Med-DualLoRA under the MIT License. The version used for the MICCAI 2026 experiments is archived under the GitHub tag `miccai2026` to ensure reproducibility. 

\paragraph{\textbf{Evaluation Metrics}} All trained models are evaluated on the original external test set of each dataset. For disease detection, and given the class imbalance, we report balanced accuracy, sensitivity, specificity, and F1-score per client. Moreover, we compute a sample-size–weighted average across clients for both centralized and FL settings, consistent with the FedAvg aggregation protocol. Client-specific models are trained and evaluated on their local dataset. Statistical significance across experimental scenarios is assessed using paired t-tests.

\paragraph{\textbf{Communication Budget \& LoRA Blocks Analysis.}}
To quantify the trade-off between predictive performance and communication cost, we conduct a structured ablation study varying the number of transformer blocks equipped with LoRA adapters. Since the backbone of the FM remains frozen, only the trainable components are transmitted during federated aggregation. Thus, the communication overhead scales with the number of adapted blocks and the rank $r$. 

For a linear layer $W \in \mathbb{R}^{d \times m}$ equipped with LoRA of rank $r$, the number of transmitted parameters scales linearly with the layer dimensions and rank. When LoRA is applied to $L$ transformer blocks and $P$ projections per block, the per-round communication cost is $ \text{Params}_{\text{comm}} = L \cdot P \cdot r (d + m)$. The communication budget in MB is computed by multiplying the number of transmitted parameters by their storage size (32-bit floating point in our experiments). 


\input{tables/table_main_results}
\input{figures/fig_performance_pareto}

\section{Results}
\paragraph{\textbf{Centralized training provides an upper bound.}}
As expected, centralized training yields the strongest overall performance (Table~\ref{tab:results_baselines_grouped}). Full fine-tuning achieves a weighted balanced accuracy of 0.745 ± 0.049, while centralized head-only fine-tuning reaches 0.810 ± 0.124. LoRA in the centralized setting performs comparably (0.780 ± 0.042), demonstrating that PEFT can approximate full fine-tuning when data is pooled without statistically significant differences. Across sites, sensitivity is consistently high ($>0.94$), whereas specificity varies substantially, indicating a bias toward the diseased (majority) class. 

\paragraph{\textbf{Strong inter-client variability in local models.}} Client-specific models trained on local datasets show a drop in performance with respect to data centralized models (Fig.~\ref{fig:performance-pareto}a). Smaller datasets such as Canon exhibit low balanced accuracy (0.611 ± 0.096) and very low specificity (0.222 ± 0.192), suggesting majority-class bias due to limited sample size. In contrast, larger datasets such as GE achieve substantially higher performance (balanced accuracy 0.756 ± 0.067). The variability observed across vendors highlights the statistical heterogeneity.

\paragraph{\textbf{Standard PEFT degrades in FL.}} When moving to FL, performance drops significantly for PEFT baselines  (Table~\ref{tab:results_baselines_grouped}). Federated head-only fine-tuning collapses to a balanced accuracy of 0.514 ± 0.024, driven by near-perfect sensitivity (0.997 ± 0.005), but extremely low specificity (0.03 ± 0.053), indicating that it suffers from severe class imbalance and insufficient model capacity. Federated LoRA improves performance (0.693 ± 0.032 balanced accuracy) but remains below its centralized counterpart, although the drop is not statistically significant. These results suggest that reducing trainable parameters alone is insufficient in heterogeneous FL, where local updates are data-limited and statistically misaligned across clients.

\paragraph{\textbf{Med-DualLoRA improves robustness in FL.}} Our proposed Med-DualLoRA methodology achieves a balanced accuracy of 0.768 ± 0.026 in the FL setting, offering a statistically significant improvement over federated LoRA (+7.5\%; p=0.0371) and head-only fine-tuning (+25.4\%; p<0.001). Importantly, Med-DualLoRA restores specificity (0.612 vs. 0.417 for LoRA and 0.03 for head-only) while maintaining high sensitivity (0.924), leading to balanced performance across classes. Variance across runs remains low (std 0.026 for balanced accuracy), indicating stable optimization. Although the F1-score for LoRA is numerically higher, this difference is not statistically significant. Notably, Med-DualLoRA performs comparably to centralized fine-tuning (p>0.05), effectively closing the federated–centralized performance gap. Furthermore, as illustrated in Fig.~\ref{fig:performance-pareto}a, Med-DualLoRA maintains competitive performance across vendors, including smaller datasets, supporting the hypothesis that additive separation of global and local low-rank components mitigates cross-client interference.

\paragraph{\textbf{Communication-accuracy trade-off.}} Pareto frontier analysis reveals the trade-off between communication cost per FL round and predictive performance (Fig.~\ref{fig:performance-pareto}b). Head-only adaptation provides the minimal communication baseline (3 MB/round) but lies at the lowest accuracy point (0.514 ± 0.024). Med-DualLoRA with all transformer blocks (12) achieves the highest balanced accuracy (0.768 ± 0.026) at $\sim$154 MB/round and dominates standard LoRA at the same communication cost. Notably, applying Med-DualLoRA to only two transformer blocks ($\sim$28 MB/round) already reaches near-optimal performance (0.7465 ± 0.0178), placing it on the efficiency frontier. Intermediate configurations (4–10 blocks) lie below the frontier, indicating diminishing returns. Overall, Med-DualLoRA offers the most favorable balance between robustness and communication efficiency in heterogeneous FL.

\section{Conclusions and Discussion}
While centralized training provides an optimistic upper bound, it is often impractical in real-world. Simple PEFT strategies tend to degrade under heterogeneous, non-IID client distributions. Med-DualLoRA addresses this limitation by disentangling shared and local low-rank adaptations, maintaining performance with respect to centralized and reducing communication overhead. Notably, we proved that adapting only a small subset of transformer blocks already approaches peak performance, underscoring the efficiency of the proposed design in small and imbalanced local datasets.

Performance variability across vendors reflects realistic domain shift and limited minority-class representation, particularly in small cohorts where metric instability is expected. We preserved the original ACDC and M\&Ms data splits to ensure comparability with prior work, although this may have amplified client-level class imbalance, suggesting stratified partitioning as future work. Although FedAvg enables collaborative learning, it favors dominant distributions under non-IID settings~\cite{ref_mcmahan2017fedavg}. Integrating Med-DualLoRA with advanced aggregation strategies such as FedProx~\cite{ref_li_fedprox}, Scaffold~\cite{ref_scaffold}, or the Rate-My-LoRA approach~\cite{ref_he_ratemylora} may further improve performance and avoid site-specific bias. Finally, broader comparisons with emerging PEFT methods beyond LoRA, such as AdapterFusion~\cite{pfeiffer2020adapterfusion} or vision prompt tuning~\cite{ref_jia2022visual}, and extension to recent multimodal CMR FMs~\cite{ref_zhang_vita} remain promising directions for future work.

Overall, Med-DualLoRA demonstrates that structurally disentangling shared and local adaptations enables effective, client-aware, and communication-efficient federated fine-tuning of 3D CMR FMs, supporting scalable and privacy-preserving deployment across heterogeneous clinical environments.

    

\begin{credits}
\subsubsection{\ackname} 
This work is part of the project TrustAI-ES (PID2023-146751OA-I00), funded by MICIU/AEI/10.13039/501100011033.

\subsubsection{\discintname}
None.
\end{credits}

%
%
%
\bibliographystyle{splncs04}
\bibliography{my_bibliography}

\end{document}

%% file: figures/figure_federated_dual_lora.tex
\begin{figure}[t]
\centering
\includegraphics[width=\textwidth, trim={0cm 0.5cm 0cm 0cm}, clip]{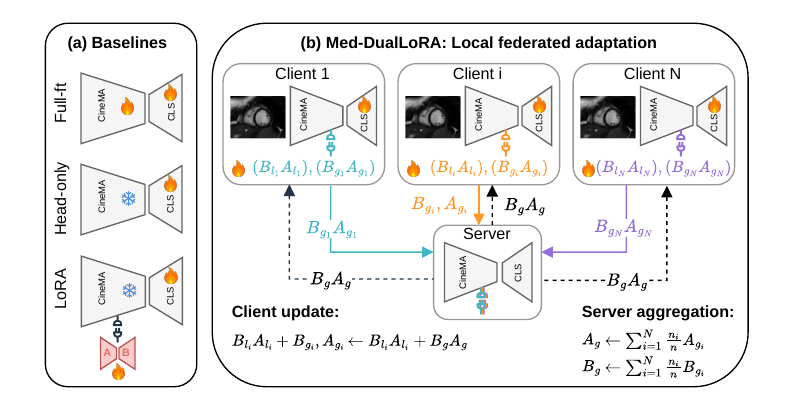}
\caption{\textbf{(a) Fine-tuning baselines. (b) Our Med-DualLoRA} method is a dual low-rank adaptation framework for collaborative learning of transferable global representations with reduced communication, while preserving client-specific differences under heterogeneous data. CLS: classifier head.}
\label{fig:dual_lora_architecture}
\end{figure}

%% file: tables/table_data_distribution.tex
\newcolumntype{L}[1]{>{\raggedright\arraybackslash}p{#1}}
\newcolumntype{C}[1]{>{\centering\arraybackslash}p{#1}}

\begin{table}[t]
\centering
\caption{Distribution of diseased (DIS) and non-diseased (NOR) samples across clients and splits, including overall totals.}
\label{tab:data_distribution}
\small
\begin{tabular}{L{0.07\columnwidth}
C{0.06\columnwidth} C{0.06\columnwidth}
C{0.06\columnwidth} C{0.06\columnwidth}
C{0.06\columnwidth} C{0.06\columnwidth}
C{0.06\columnwidth} C{0.06\columnwidth}
C{0.06\columnwidth} C{0.06\columnwidth}
C{0.06\columnwidth} C{0.06\columnwidth}}
\hline
 & \multicolumn{2}{c}{Canon} 
 & \multicolumn{2}{c}{GE} 
 & \multicolumn{2}{c}{Philips} 
 & \multicolumn{2}{c}{Siemens} 
 & \multicolumn{2}{c}{ACDC}
 & \multicolumn{2}{c}{\textbf{Total}} \\
 & DIS & NOR 
 & DIS & NOR 
 & DIS & NOR 
 & DIS & NOR 
 & DIS & NOR
 & DIS & NOR \\
 \hline
 Train & 28 & 7 & 53 & 20  & 78 & 26  & 151 & 16 & 72 & 18 & 382 & 87 \\
 Val & 3  & 4 & 14 & 1 & 15 & 7 & 32 & 3 & 8 & 2 & 72 & 17 \\
 Test & 5  & 3 & 15 & 2 & 19 & 4 & 34 & 3 & 40 & 10 & 113 & 22 \\
 \hline
 \textbf{Total} & 36 & 14 & 82 & 23 & 112 & 37 & 217 & 22 & 120 & 30 & 567 & 126 \\  
 \hline
\end{tabular}
\end{table}

%% file: tables/table_main_results.tex
\begin{table}[t]
\centering
\caption{Performance metrics (mean ± std) for centralized and federated baselines, and Med-DualLoRA. Asterisks indicate statistically significant differences compared to Med-DualLoRA using paired t-test: $^{*}$ p<0.05, $^{**}$ p<0.01.}
\label{tab:results_baselines_grouped}
\small
\begin{tabular}{l @{\hskip 3mm} c @{\hskip 3mm} c @{\hskip 3mm} c @{\hskip 3mm} c}
\hline
\textbf{Experiment} & \textbf{Balanced Acc.} & \textbf{Sensitivity} & \textbf{Specificity} & \textbf{F1} \\
\hline
\multicolumn{5}{l}{\textbf{Centralized}} \\
Full-ft                        & 0.745 ± 0.049$^{~~~}$ & 0.953 ± 0.013$^{~~}$ & 0.538 ± 0.111$^{~~}$ & 0.927 ± 0.006 \\
Head-only                      & 0.810 ± 0.124$^{~~~}$ & 0.949 ± 0.006$^{~~}$ & 0.672 ± 0.250$^{~~}$ & 0.937 ± 0.022 \\
LoRA                            & 0.780 ± 0.042$^{~~~}$ & 0.943 ± 0.017$^{~~}$ & 0.616 ± 0.092$^{~~}$ & 0.929 ± 0.009 \\
\hline
\multicolumn{5}{l}{\textbf{Federated}} \\
Head-only                      & 0.514 ± 0.024$^{**}$ & 0.997 ± 0.005$^{*}$ & 0.030 ± 0.053$^{**}$ & 0.909 ± 0.000 \\
LoRA                            & 0.693 ± 0.032$^{*~}$  & 0.970 ± 0.005$^{~}$       & 0.417 ± 0.061$^{*~}$  & 0.927 ± 0.006 \\
Med-Dual LoRA                   & 0.768 ± 0.026 $^{~}$       & 0.924 ± 0.020$^{~}$       & 0.612 ± 0.039 $^{~}$       & 0.921 ± 0.009 \\
\hline
\end{tabular}
\end{table}

%% file: figures/fig_performance_pareto.tex
\begin{figure}[t]
\centering
\includegraphics[width=\textwidth, trim={0.6cm 0.45cm 0.6cm 0.3cm}, clip]{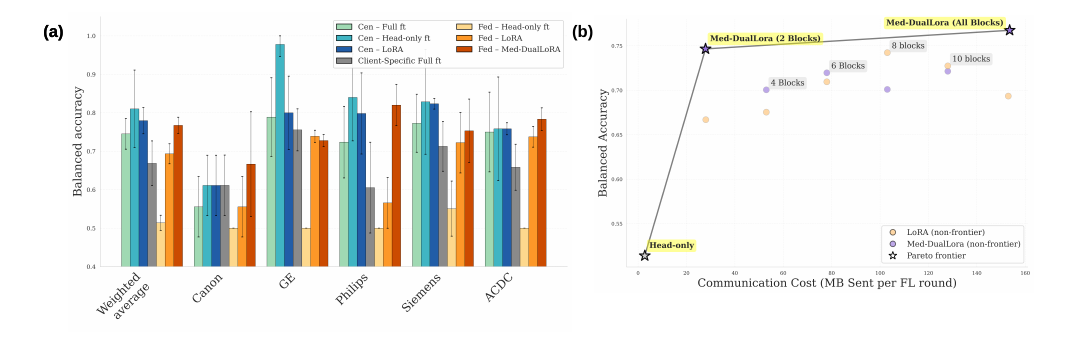}
\caption{\textbf{(a) Client-wise performance under centralized, client-specific, and federated training.} Model performance is shown per client and averaged across clients (size-normalized); error bars indicate mean ± standard deviation across runs.  \textbf{(b) Pareto frontier of communication cost versus balanced accuracy.} Stars denote frontier configurations, illustrating the trade-off between performance and communication per round.
}
\label{fig:performance-pareto}
\end{figure}
